\documentclass[conference]{IEEEtran}
\IEEEoverridecommandlockouts
\usepackage{cite}
\usepackage[numbers,sort&compress]{natbib}

\usepackage[utf8]{inputenc}
\usepackage{amsmath,amssymb}
\usepackage{multirow}
\usepackage{graphicx}
\usepackage{bm}
\usepackage{enumerate}
\usepackage{comment}
\usepackage{xcolor}
\usepackage{url}
\usepackage{color,soul}
\usepackage{subcaption}
\usepackage{float}
\usepackage{booktabs} % For prettier tables
\usepackage{csquotes}

\definecolor{light-gray}{gray}{0.8}

\usepackage{amsmath,amssymb,amsfonts}
\usepackage{algorithmic}
\usepackage{graphicx}
\usepackage{textcomp}
\usepackage{xcolor}
\usepackage{subcaption}

\def\BibTeX{{\rm B\kern-.05em{\sc i\kern-.025em b}\kern-.08em
    T\kern-.1667em\lower.7ex\hbox{E}\kern-.125emX}}

\makeatletter
\newcommand{\linebreakand}{%
  \end{@IEEEauthorhalign}
  \hfill\mbox{}\par
  \mbox{}\hfill\begin{@IEEEauthorhalign}
}
\makeatother

\begin{document}

\title{Enhancing Exchange Rate Forecasting with Explainable Deep Learning Models\\}

\author{

\small % Set font size to 10pt

\begin{tabular}[t]{c@{\extracolsep{8em}}c} 

1\textsuperscript{st} Shuchen Meng\textsuperscript{*} & 2\textsuperscript{nd} Andi Chen \\
\textit{Central University of Finance and Economics, Beijing, China} & \textit{Independent Researcher, Beijing, China} \\
scmeng19@163.com & \\

\\

3\textsuperscript{rd} Chihang Wang & 4\textsuperscript{th} Mengyao Zheng \\
\textit{New York University, New York, USA} & \textit{Harvard University, Boston, USA} \\

\\

5\textsuperscript{th} Fangyu Wu & 6\textsuperscript{th} Xupeng Chen \\
\textit{IEEE Student Member, New York, USA} & \textit{New York University, New York, USA} \\

\\

7\textsuperscript{th} Haowei Ni & 8\textsuperscript{th} Panfeng Li \\
\textit{Columbia University, New York, USA} & \textit{University of Michigan, Ann Arbor, USA} \\

\end{tabular}
}

\maketitle

\begin{abstract}
Accurate exchange rate prediction is fundamental to financial stability and international trade, positioning it as a critical focus in economic and financial research. Traditional forecasting models often falter when addressing the inherent complexities and non-linearities of exchange rate data. This study explores the application of advanced deep learning models, including LSTM, CNN, and transformer-based architectures, to enhance the predictive accuracy of the RMB/USD exchange rate. Utilizing 40 features across 6 categories, the analysis identifies TSMixer as the most effective model for this task. A rigorous feature selection process emphasizes the inclusion of key economic indicators, such as China-U.S. trade volumes and exchange rates of other major currencies like the euro-RMB and yen-dollar pairs. The integration of grad-CAM visualization techniques further enhances model interpretability, allowing for clearer identification of the most influential features and bolstering the credibility of the predictions. These findings underscore the pivotal role of fundamental economic data in exchange rate forecasting and highlight the substantial potential of machine learning models to deliver more accurate and reliable predictions, thereby serving as a valuable tool for financial analysis and decision-making.
\end{abstract}


\section{Introduction}

Since the dissolution of the Bretton Woods system, the adoption of a floating exchange rate regime has introduced significant challenges in risk management for market participants. The volatility of the RMB/USD exchange rate, particularly during periods of trade tensions, has heightened the uncertainty faced by those engaged in the foreign exchange market. The People's Bank of China's reform of the exchange rate fixing mechanism on August 11, 2015, further increased the marketization of the RMB exchange rate, leading to greater exchange rate volatility and increased foreign exchange risk. This has underscored the critical need for accurate exchange rate forecasting and effective risk management strategies.

China’s growing role in the global supply chain, especially after its accession to the World Trade Organization (WTO), and the deepening economic ties between China and the United States, have made the RMB/USD exchange rate a focal point of global economic stability. Debates over the valuation of the RMB, particularly during periods of significant trade surpluses with the U.S., have led to multiple rounds of discussions on exchange rate policy. The RMB exchange rate reform in 2005 and the subsequent rounds of monetary policy adjustments by the Federal Reserve, especially during the 2007 financial crisis, further complicated the dynamics of the RMB/USD exchange rate. The “8.11 Exchange Rate Reform” in 2015, which introduced a more flexible exchange rate mechanism, and the intensified trade frictions since 2018, have put additional depreciation pressure on the RMB, making accurate forecasting of the RMB/USD exchange rate increasingly important~\cite{zhang2001time,galeshchuk2016neural,rossi2013exchange}.

Previous research on exchange rate forecasting has primarily focused on theoretical and quantitative models. Theoretical models often emphasize the equilibrium state of exchange rates, which can be difficult to achieve or maintain in practice, making short- to medium-term predictions particularly challenging. Quantitative models focus on the exchange rate's own dynamics while often neglecting other critical influencing factors. Moreover, these models have struggled to produce consistent results across different studies.

In recent years, there has been a notable shift towards using big data approaches in forecasting models, bypassing the need for complex mathematical modeling and allowing for more flexible model forms without predefined structures. The inherent complexity and non-linearity of exchange rate data have led to the application of non-linear methods, such as chaos theory, non-parametric methods, and machine learning techniques, which have shown potential to improve forecasting accuracy. Studies like those of LeBaron and others have demonstrated that methods such as kernel ridge regression can significantly enhance the prediction of financial volatility, although some researchers, such as Mourer, have found that these methods do not always outperform simple autoregressive models in all contexts.

Since the end of the Bretton Woods system, the transition to a floating exchange rate regime has posed significant challenges for managing risk, especially regarding the RMB/USD exchange rate. China’s integration into the global economy post-WTO accession, along with its deepening economic ties with the U.S., has made this exchange rate crucial for global economic stability. The 2015 reform of China’s exchange rate mechanism increased market-driven fluctuations, further intensified by trade tensions, which have heightened volatility. This evolving landscape has led to debates over the RMB’s valuation and spurred numerous policy discussions. Consequently, precise and reliable exchange rate forecasting has become essential for effective risk management in this volatile environment.

Traditional exchange rate forecasting models, both theoretical and quantitative, have struggled with consistency and often neglect critical influencing factors. For example, theoretical models may fail to account for the real-time impact of policy changes and global economic shifts, while quantitative models may not fully capture the non-linear dynamics and intricate interactions of the variables involved. Recently, there has been a shift toward big data and deep learning approaches~\cite{wang2024tssurvey}, which offer flexibility and improved accuracy in handling the complex, non-linear nature of exchange rate data. While some methods, like kernel ridge regression, have shown promise, their performance varies across different contexts, highlighting the ongoing challenges in exchange rate prediction.

Machine learning models, particularly deep learning models, have increasingly been applied to predicting time series and economic variables. Despite their advantages in handling complex, non-linear data without requiring explicit assumptions about the underlying data distribution, these models are often criticized for their \enquote{black box} nature and lack of interpretability. Recent advancements, such as the application of Grad-CAM~\cite{selvaraju2020grad} and attention mechanisms, have begun to address these issues, making it possible to visualize model predictions and understand the underlying decision-making processes. However, applying these interpretability techniques has been mostly limited to fields like image recognition and natural language processing, with relatively few studies applying them to economic forecasting.

Given the challenges of traditional models and the potential of machine learning approaches, this study seeks to explore the use of advanced deep learning models, including CNNs, RNNs, MLPs~\cite{li2021survey,sherstinsky2020fundamentals,popescu2009multilayer}, and transformer-based~\cite{transformer2017} architectures, for predicting the RMB/USD exchange rate. By incorporating a comprehensive set of features—drawn from economic indicators, trade data, and other currency pairs—and employing advanced feature selection techniques, this research aims to enhance predictive accuracy, identify the most relevant factors influencing exchange rate fluctuations, and enhance the interpretability of the model predictions.

\paragraph{Contributions of This Study}

\begin{itemize}
    \item  \textbf{Application of Deep Learning Models:} This study provides an initial analysis of the effectiveness of deep learning models in exchange rate prediction, using MSE and MAE as key metrics to identify the best-performing models.
    \item  \textbf{Enhancement of Predictive Performance:} To improve the accuracy of machine learning models, this study employs various techniques, including feature selection, to reduce redundancy and retain the most relevant subset of features for exchange rate forecasting.
    \item  \textbf{Analysis of Influential Factors Over Time:} By applying attention mechanisms, this study enhances the interpretability of machine learning models, offering insights into how different factors influence exchange rate predictions across different periods. This analysis aims to uncover which aspects of economic data the models prioritize during the prediction process, thereby providing a more nuanced understanding of the underlying dynamics.
\end{itemize}

\section{Methods}
\subsection{Data Collection and Preprocessing}

The data selected for this study is derived from three sources:

\textbf{Macroeconomic Statistics: }This includes key indicators like import/export values and short-term capital flows. While critical for decision-making, these statistics often suffer from delays, inaccuracies, and lack of predictive power. Branson’s (1975) asset portfolio approach to exchange rates, integrating purchasing power parity and risk-return models, is hampered by such data lags.  \textbf{Financial Market Trading Data: }Includes real-time data like stock indices, concept stock indices, and exchange rates. These data are timely, transparent, and reflect market expectations, making them valuable for exchange rate predictions. 
 \textbf{Macroeconomic Variables:} Covers indicators like M2 (monetary policy), spot interest rates, and price indices (inflation). Most data are daily; monthly data are treated as daily. This practice reflects real-world usage, where outdated data often guide predictions. Building upon recent advancements in financial data processing, we incorporated techniques inspired by Chang's research to significantly enhance our model's processing efficiency, particularly in handling diverse data frequencies and potential redundancies.
 
We summarize the data we collected in Table. \ref{table2}. All data were sourced from the WIND Financial Terminal. Variables include the RMB/USD exchange rate and other relevant economic and financial indicators. Traditional regression models would struggle with multicollinearity and parameter estimation due to the extensive set of variables. However, machine learning models mitigate these issues. Post-2015, the volatility of the USD/RMB exchange rate increased, leading us to focus on this period. We applied z-score normalization to stabilize the model training process.

% \begin{table*}[ht]
% \centering
% \caption{Factors Description Table}
% \resizebox{0.9\textwidth}{!}{%
% \begin{tabular}{p{5cm}p{8cm}l}
% \toprule
% Indicator & Description & Indicator Name \\ \midrule
% Exchange Rate & Offshore Spot RMB Closing Price & \textbf{rate} \\ 

% Long-term Bond Yield Differential (U.S.) & 10-Year Government Bond Yield - U.S. Federal Funds Rate & \textbf{udr} \\ 
% CPI (U.S.) & U.S. Consumer Price Index & \textbf{cpiu} \\ 

% CSI 300 Index & CSI 300 Index & \textbf{HS300} \\ 

% Import and Export Value & Current Month Import and Export Value & \textbf{trade} \\ 
% Import Value & Current Month Import Value & \textbf{inputu} \\  
% M2 (U.S.) & M2 in the United States & \textbf{m2u2} \\  
% EUR/USD & EUR/USD Exchange Rate & \textbf{EURUSD} \\ 
% AUD/USD & AUD/USD Exchange Rate & \textbf{AUDUSD} \\ 
% USDX & U.S. Dollar Index Published by ICE & \textbf{USDX} \\ \bottomrule
% \end{tabular}
% }
% \label{table1}
% \end{table*}

\begin{table*}[ht]
\centering
\caption{Factors Description Table}
\resizebox{0.9\textwidth}{!}{%
\begin{tabular}{lp{5cm}p{8cm}l}
\toprule
Category & Indicator & Description & Indicator Name \\ \midrule
\multirow{9}{*}{Fundamental Data} 
& Exchange Rate & Offshore Spot RMB Closing Price & \textbf{rate} \\ 
& Interest Rate & U.S. Federal Funds Rate & rusa \\ 
& Interest Rate & PBoC Benchmark Deposit Rate & rchn \\ 
& Interest Rate Differential & Interest Rate Differential Between Two Countries & dr \\ 
& Long-term Bond Yield Differential (China) & 10-Year Government Bond Yield - Bank Lending Rate & ydr \\ 
& Long-term Bond Yield Differential (U.S.) & 10-Year Government Bond Yield - U.S. Federal Funds Rate & \textbf{udr} \\ 
& CPI (U.S.) & U.S. Consumer Price Index & \textbf{cpiu} \\ 
& CPI (China) & China Consumer Price Index & cpic \\ 
& Consumer Confidence Index & China Consumer Confidence Index & ccp \\ \midrule

\multirow{6}{*}{Stock Index Data} 
& Dow Jones Index & Dow Jones Index & dowjones \\ 
& MSCI China Index & MSCI China Index & MSCIAAshare \\ 
& CSI 300 Index & CSI 300 Index & \textbf{HS300} \\ 
& S\&P 500 China Index & S\&P 500 China Index & sprd30.ci \\ 
& Nasdaq China Index & Nasdaq China Index & nyseche \\ 
& Hang Seng China Enterprises Index & Hang Seng China Enterprises Index & hscei.hi \\ \midrule

\multirow{3}{*}{Current Account} 
& Import and Export Value & Current Month Import and Export Value & \textbf{trade} \\ 
& Export Value & Current Month Export Value & output \\ 
& Import Value & Current Month Import Value & \textbf{inputu} \\ \midrule

\multirow{5}{*}{Capital Account} 
& FDI & Foreign Direct Investment & fdix \\ 
& FDI & Actual Foreign Investment in the Current Month & fdi \\ 
& PI & Portfolio Investment & PI \\ 
& OI & Other Investment & OI \\ 
& Short-term International Capital Flow & Foreign Exchange Reserves Change in the Current Month \newline Net Current Account Balance & cf \\
\midrule
\multirow{3}{*}{Currency Market} 
& M2 (China) & M2 in China & cm2 \\ 
& M2 Growth (China) & M2 Growth in China & dm2 \\ 
& M2 (U.S.) & M2 in the United States & \textbf{um2} \\ \midrule

\multirow{6}{*}{Exchange Rates} 
& USD/JPY & USD/JPY Exchange Rate & USDJPY \\ 
& GBP/USD & GBP/USD Exchange Rate & GBPUSD \\ 
& EUR/CHN & EUR/CHN Exchange Rate & EURCHN \\ 
& EUR/USD & EUR/USD Exchange Rate & \textbf{EURUSD} \\ 
& AUD/USD & AUD/USD Exchange Rate & \textbf{AUDUSD} \\ 
& USD/CAD & USD/CAD Exchange Rate & USDCAD \\ 
& USD/CHF & USD/CHF Exchange Rate & USDCHF \\ 
\midrule
\multirow{1}{*}{U.S. Dollar Index} 
& USDX & U.S. Dollar Index Published by ICE & \textbf{USDX} \\ \bottomrule
\end{tabular}
}
\label{table2}
\end{table*}

\subsection{Feature Selection}

In the data introduction, we identified three sources of data in table \ref{table2}: macroeconomic statistics (e.g., import/export figures, capital flows), financial market data (e.g., stock indices, real-time exchange rates), and macroeconomic variables reflecting economic fundamentals (e.g., M2, spot interest rates, price indices).

However, the selection of these features posed challenges. The data varied in frequency—mostly daily, but some monthly, requiring monthly values to be treated as daily values. This reflects how market participants often rely on past monthly data when real-time information is unavailable. Additionally, there was potential redundancy among features, such as between major currency exchange rates and the U.S. dollar index or between different stock indices. With 40 features across six categories, we initially conducted a preliminary analysis by predicting exchange rates using individual features to assess their importance. The results were underwhelming, with no single feature proving particularly significant.

To improve model accuracy, we performed feature selection using a ridge regression-based wrapper method, focusing on the subset of features that best enhanced the performance of the final learning model. We selected the following features as the final feature set for deep learning forecasting models: \textbf{HS300, cpiu, audusd, eurusd, m2u2, inputu, trade, udr, usdx, date.}

\subsection{Models}

We leverage the recently developed deep learning models, including TSMixer \cite{tsmixer2023}, FEDformer
\cite{zhou2022fedformer}, LSTM \cite{lstm1997}, PatchTST \cite{nie2022time}, TimesNet \cite{wu2022timesnet}, Transformer \cite{transformer2017}, MLP \cite{mlp1989}, TCN \cite{tcn2018}, and iTransformer \cite{chang-trans-24} for exchange rate prediction.

\textbf{LSTM}: We used three layers of LSTM with hidden dimensions of 32, 64, and 64. \textbf{TCN}: Applied in multi-variate time series prediction, using three layers of dilated temporal convolutions with residual connections. Kernel sizes were (3, 3, 5) with a dilation rate of (2, 2, 2). \textbf{TSMixer}: An MLP-based neural network with 5 layers, each capturing and mixing temporal features. Configured with a maximum feature dimension of 16 for balanced learning and efficiency. \textbf{TimesNet}: Combines convolutional layers to model local and global temporal patterns, transforming 1D time series into 2D tensors. The architecture includes 1 FFT block and 4 convolutional blocks. \textbf{PatchTST}: Divides input sequences into patches, processed independently using 6 transformer layers with 8 heads per layer, and a patch size of 12. \textbf{iTransformer}: Trained with improved positional encoding and attention mechanisms, focusing on feature dimension attention. Consists of 4 transformer blocks, each with 2 layers and 8 attention heads. \textbf{Fedformer}: Configured with 3 encoder and 2 decoder layers, each using 8-head self-attention. The feature dimension is 256, with feedforward networks at 512. \textbf{Transformer}: Implemented with 6 layers in both encoder and decoder, using 8-head self-attention and a feature dimension of 512. Each layer has a feedforward network with an inner dimension of 2048, with positional encoding to capture temporal dependencies. \textbf{MLP}: Configured as a fully connected network with 3 hidden layers, each containing 128 neurons and ReLU activation for non-linearity.

All the models were trained using the MAE loss function. We used the Adam optimizer with the following hyperparameters: learning rate (\( \text{lr} \)) = \( 10^{-3} \), \( \beta_1 \) = 0.9, \( \beta_2 \) = 0.999. The models were trained for 1000 epochs with five-fold cross-validation.

\section{Experiments}

 We tested nine deep-learning models on our data. We tried input and prediction length pairs: (32, 16), (48, 24), (64, 32), (96, 48), and (128, 64).

\subsection{Evaluation Metrics}

In this study, model performance was evaluated using two key metrics: Mean Squared Error (MSE) and Mean Absolute Error (MAE). These metrics were selected to provide a thorough evaluation of the model's predictive accuracy by quantifying the differences between the predicted and actual values. MSE emphasizes larger errors by squaring them, while MAE offers a straightforward measurement of the average error magnitude. To ensure the reliability of the results, the metrics for each model were averaged across five-fold cross-validation, offering a more robust performance assessment.

\subsection{Performance Analysis}

\begin{table*}[htbp]
\centering
\caption{Performance Metrics of Different Models with Different Prediction Lengths}
\resizebox{0.75\textwidth}{!}{%
\begin{tabular}{l|r|r|r|r|r|r|r|r|r|r}
\toprule
 Model & \multicolumn{5}{c|}{MAE} & \multicolumn{5}{c}{MSE} \\
 \midrule
Prediction Length & 16 & 24 & 32 & 48 & 64 & 16 & 24 & 32 & 48 & 64 \\
\midrule
\textbf{TSMixer} & \textbf{0.032} & \textbf{0.039} & \textbf{0.045} & \textbf{0.054} & \textbf{0.063} & \textbf{0.002} & \textbf{0.003} & \textbf{0.004} & \textbf{0.006} & \textbf{0.007} \\
\textbf{FEDformer} & 0.033 & 0.042 & 0.049 & 0.063 & 0.086 & 0.002 & 0.004 & 0.004 & 0.007 & 0.012 \\
\textbf{iTransformer} & 0.035 & 0.054 & 0.072 & 0.089 & 0.064 & 0.002 & 0.005 & 0.008 & 0.011 & 0.008 \\
%\textbf{DLinear} & 0.035 & 0.042 & 0.046 & 0.055 & 0.064 & 0.002 & \textbf{0.003} & 0.004 & 0.006 & 0.008 \\
\textbf{PatchTST} & 0.039 & 0.060 & 0.056 & 0.082 & 0.125 & 0.003 & 0.011 & 0.006 & 0.012 & 0.034 \\
\textbf{TimesNet} & 0.038 & 0.052 & 0.063 & 0.084 & 0.104 & 0.003 & 0.005 & 0.008 & 0.012 & 0.020 \\
\textbf{Transformer} & 0.042 & 0.057 & 0.067 & 0.088 & 0.095 & 0.004 & 0.007 & 0.008 & 0.014 & 0.015 \\
\textbf{MLP} & 0.053 & 0.074 & 0.124 & 0.103 & 0.233 & 0.006 & 0.013 & 0.043 & 0.017 & 0.108 \\
\textbf{TCN} & 0.049 & 0.080 & 0.149 & 0.128 & 0.063 & 0.004 & 0.009 & 0.026 & 0.020 & 0.007 \\
\textbf{LSTM} & 0.047 & 0.056 & 0.064 & 0.098 & 0.132 & 0.005 & 0.006 & 0.007 & 0.015 & 0.027 \\
\bottomrule
\end{tabular}
}
\label{table:metrics}
\end{table*}

The overall performance of the evaluated models, as presented in Table \ref{table:metrics}, highlights the advantage of modern transformer-based architectures over traditional deep learning models (MLP, TCN, and LSTM) in the task of time series forecasting. Notably, TSMixer, FEDformer, and iTransformer consistently achieved superior results, demonstrating their ability to accurately capture and predict complex temporal patterns.

Among all models, TSMixer has the lowest MAE and MSE values across multiple prediction lengths. This performance underscores the model’s effectiveness in handling diverse temporal dependencies. It is highly effective in both long-term forecasting benchmarks for multivariate time series and real-world forecasting tasks, which in our case, the exchange rate prediction. The model's capacity to generalize well across varying forecasting horizons further solidifies its position as the top performer.

% FEDformer closely follows TSMixer, with strong performance across both metrics. The model's architecture, which leverages Fourier Enhanced Decomposed blocks, excels at disentangling long-term trends from short-term fluctuations. This capability allows FEDformer to produce reliable and accurate forecasts, particularly by effectively capturing both periodic patterns and transient anomalies in the data. iTransformer also significantly improved over classical models like MLP, TCN, and LSTM. The inverted self-attention mechanism in iTransformer facilitates a more nuanced understanding of the temporal and feature relationships within the data, which is especially beneficial for longer prediction horizons. This results in better overall performance compared to traditional models.

% In contrast, traditional models such as MLP and LSTM struggled to match the accuracy of the transformer-based models. The MLP model, lacking sequential modeling capabilities, showed higher error rates, particularly in the MSE metric. Although LSTM can model long-term dependencies, it was surpassed by newer transformer-based models due to its limitations in effectively capturing intricate temporal patterns across different scales. TCN performed better than MLP and LSTM but still fell short compared to TSMixer and FEDformer. While TCN’s convolutional structure allows for improved temporal understanding over MLP, it lacks the dynamic feature selection present in attention-based models, leading to less accurate forecasts.

FEDformer closely follows TSMixer, performing strongly across both metrics. Its architecture, leveraging Fourier Enhanced Decomposed blocks, excels at separating long-term trends from short-term fluctuations, capturing periodic patterns and transient anomalies for reliable forecasts. iTransformer, with its inverted self-attention mechanism, also outperforms classical models like MLP, TCN, and LSTM by providing a nuanced understanding of temporal and feature relationships, particularly for longer prediction horizons.

In contrast, traditional models like MLP and LSTM struggled to match the accuracy of transformer-based models. MLP, lacking sequential modeling, showed higher error rates, especially in MSE. While LSTM can model long-term dependencies, it was surpassed by transformer models in capturing intricate temporal patterns. TCN performed better than MLP and LSTM but still lagged behind TSMixer and FEDformer, as its convolutional structure, while improving temporal understanding, lacks the dynamic feature selection of attention-based models, resulting in less accurate forecasts.

\begin{figure}[htbp]
    \centering
    \includegraphics[width=\linewidth]{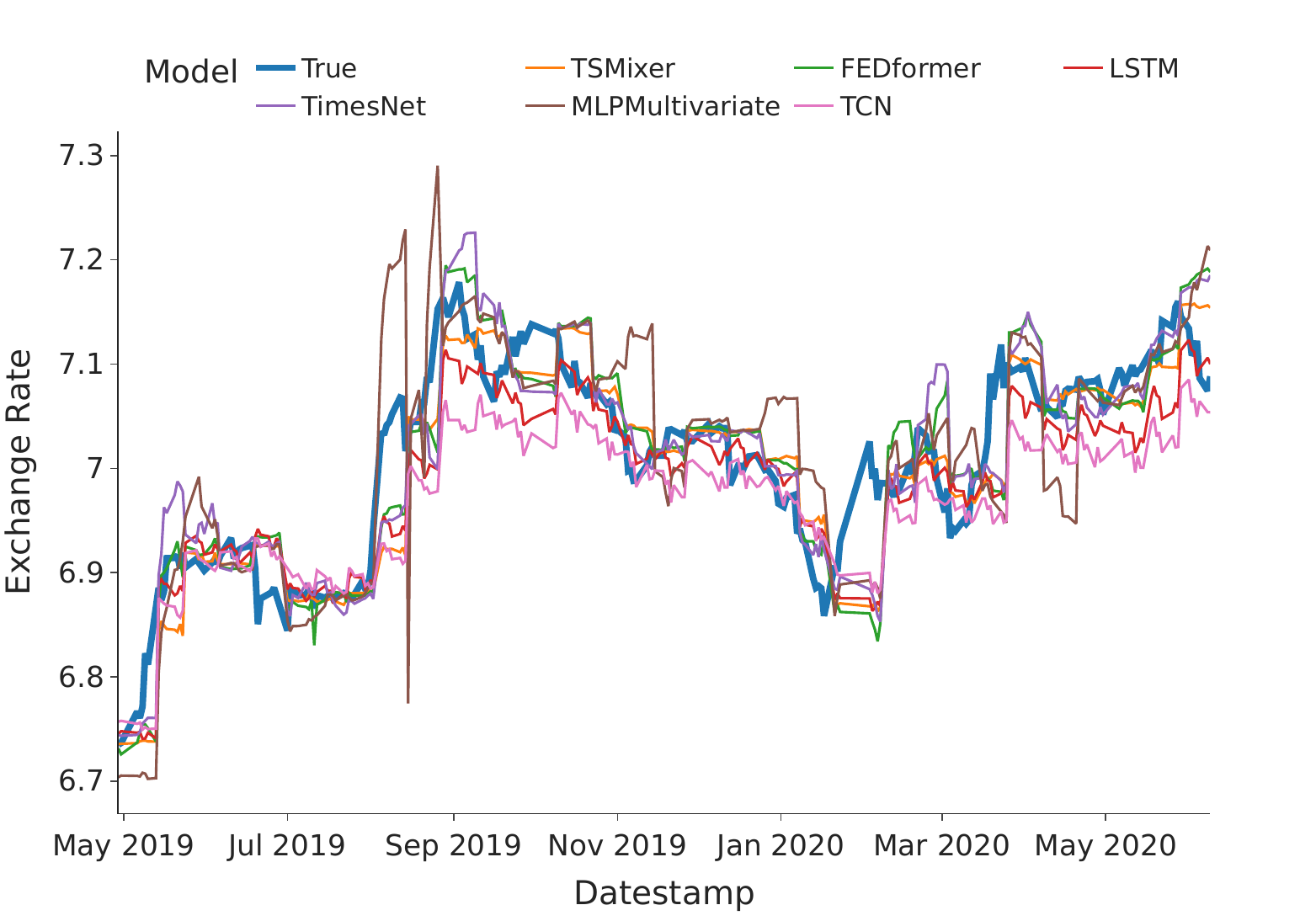}
    \caption{Comparison of True Values vs. Predicted Outputs from six models: TSMixer, FEDformer, LSTM, TimesNet, MLP, TCN}
    \label{fig:forecast_comparison}
\end{figure}

The findings are further supported by Figure \ref{fig:forecast_comparison}, where the predicted outputs are compared with the true exchange rate values. TSMixer and FEDformer closely track the actual values, reflecting their high accuracy. TCN and TimesNet also perform well but show slightly more deviation. In contrast, LSTM and MLP, despite being traditional models, exhibit the most variability.

In conclusion, the analysis confirms that TSMixer, FEDformer, and iTransformer excel in capturing complex temporal dependencies and nonlinear relationships. These models consistently achieve lower MAE and MSE values, making them more suitable for real-world forecasting where precision and robustness are critical. The shift towards transformer architectures in time series forecasting is well-justified, particularly when compared to traditional deep learning methods.

\subsection{Gradient Based Feature Importance Analysis}
\begin{figure}[htbp]
    \centering\includegraphics[width=0.85\linewidth]{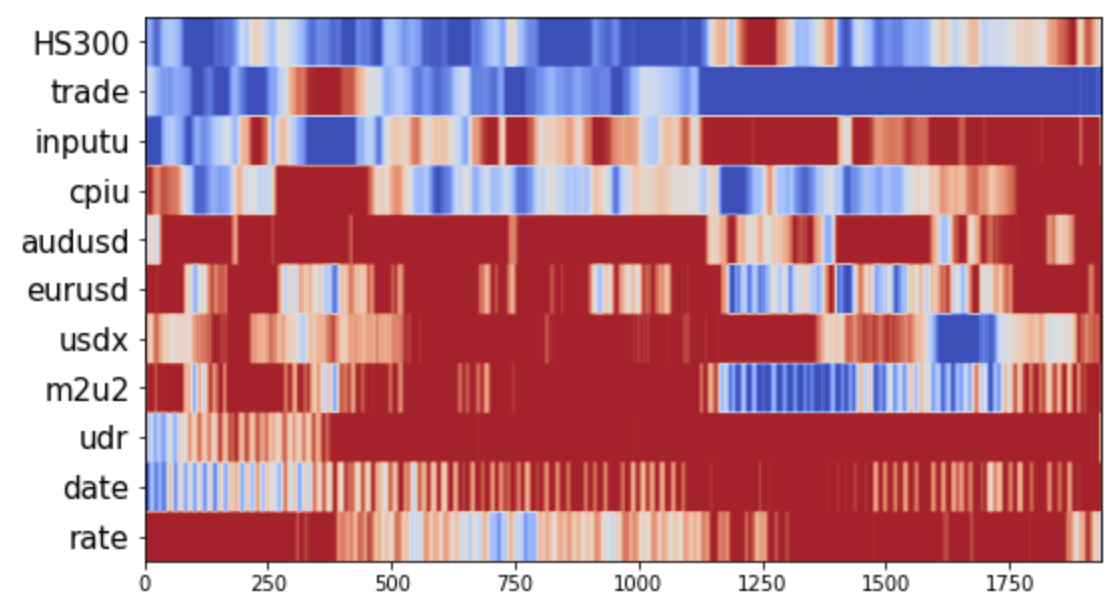}
    \caption{GradCAM visualization indicates feature contribution for forecasting.}
    \label{fig:gradcam}
\end{figure}

We applied grad-CAM~\cite{selvaraju2020grad} to obtain visual explanations for decisions from our deep learning models, making them more transparent and explainable. grad-CAM generates heatmaps that capture the TSMixer's focus during prediction on different features and time points, shown in Figure \ref{fig:gradcam}. The higher value (red) indicates the model is focusing more. We can clearly observe that the TSMixer model assigns significant weight to certain low-frequency macroeconomic variables, such as import and export volumes and trade amounts. However, the model appears to place even greater emphasis on trade-related data. It also shows a notable focus on fundamental data, such as the yield spread between U.S. 10-year Treasury bonds and the Federal Reserve’s interest rates, as well as the U.S. Consumer Price Index. The model assigns considerable importance to other currency pairs but does not prioritize stock indices like the CSI 300. This is visually evident in the figure, where these features stand out prominently. These findings suggest that variables with different frequencies play varying roles in exchange rate forecasting, but fundamental data remains crucial. This outcome also reinforces the reliability of using machine learning models for exchange rate prediction.

\section{Discussion}

\subsection{Modern Deep Learning Models on Exchange Rate Prediction}

The application of transformer-based models in exchange rate forecasting has significantly outperformed traditional approaches like MLP, TCN, and LSTM. The superior performance of TSMixer, FEDformer, and iTransformer highlights their advanced capabilities in capturing complex temporal dependencies and nonlinear relationships. TSMixer, in particular, achieves the lowest MAE and MSE across various prediction lengths, effectively prioritizing relevant time steps through self-attention. FEDformer excels in distinguishing long-term trends from short-term fluctuations using Fourier Enhanced Decomposed blocks, while iTransformer enhances temporal and feature relationship understanding, particularly for longer horizons. These models set a new benchmark in predictive accuracy for exchange rate forecasting, with the potential to revolutionize financial analysis and decision-making.

\subsection{Theoretical and Real World Implications}

The success of transformer-based models in this study suggests a paradigm shift in exchange rate forecasting, where advanced deep learning architectures increasingly outperform traditional models. This shift is crucial for financial markets, where accurate predictions are vital for risk management and strategic decision-making. These models' ability to handle complex, nonlinear data opens new research avenues in financial forecasting, potentially leading to even more sophisticated models. Practically, their enhanced predictive performance benefits traders, analysts, and policymakers by providing more accurate forecasts, mitigating exchange rate volatility risks, and contributing to financial market stability. Moreover, using explainable AI techniques like Grad-CAM improves model interpretability, making them more transparent and trustworthy for decision-makers.

This study successfully demonstrates the superiority of transformer-based models in forecasting the RMB/USD exchange rate, outperforming traditional deep learning models such as MLP, TCN, and LSTM. The advanced architectures of TSMixer, FEDformer, and iTransformer allow these models to capture complex temporal dependencies and nonlinear relationships more effectively, resulting in significantly lower MAE and MSE values. These findings underscore the transformative potential of transformer-based models in financial forecasting, particularly in the highly volatile context of exchange rates.

The visual analysis provided by Grad-CAM further corroborates the effectiveness of these models, highlighting their focus on the most influential features in the data. This interpretability is crucial for building trust in AI-driven financial models, ensuring that their predictions are not only accurate but also understandable to end-users. While this study focuses on exchange rate forecasting, the implications of these findings extend to other areas of financial analysis, where the ability to predict complex patterns and relationships is essential.

%In conclusion, the adoption of transformer-based models represents a significant advancement in the field of financial forecasting. Their superior performance in exchange rate prediction sets a new standard for accuracy and reliability, paving the way for more sophisticated and effective financial analysis tools in the future. Future research should explore the application of these models to other financial time series data and investigate their long-term performance in real-world market conditions.

\section{Limitations and Future Research}

Despite the promising results, this study has several limitations. The models were evaluated solely on the RMB/USD exchange rate, which may limit the generalizability of the findings to other currencies or financial instruments. Additionally, the models were tested using historical data and did not account for unexpected economic shocks or geopolitical events, which could impact forecast accuracy. While Grad-CAM enhances model interpretability, the complex nature of transformer models still presents challenges in fully understanding their decision-making processes. Furthermore, the study primarily addresses short to medium-term forecasting; the effectiveness of these models for long-term predictions and their adaptability to evolving market conditions remain areas for further investigation. Future research should focus on applying these models to a broader range of financial data, incorporating external factors, and evaluating their performance over extended periods.

\section{Conclusion}

This study demonstrates the superiority of transformer-based models in forecasting the RMB/USD exchange rate, surpassing traditional deep learning models like MLP, TCN, and LSTM. The advanced architectures of TSMixer, FEDformer, and iTransformer effectively capture complex temporal dependencies and nonlinear relationships, resulting in significantly lower MAE and MSE values. These findings highlight the transformative potential of transformer-based models in financial forecasting, particularly in volatile exchange rate contexts.

The use of Grad-CAM for visual analysis further validates these models, showing their focus on the most influential features, which enhances interpretability and builds trust in AI-driven financial models. While this study centers on exchange rate forecasting, the implications extend to other financial analyses requiring complex pattern prediction.

In conclusion, transformer-based models mark a significant advancement in financial forecasting, setting a new standard for accuracy and reliability. Future research should explore their application to other financial time series and assess their long-term performance in real-world market conditions.

\renewcommand{\bibfont}{\footnotesize}

\footnotesize{
\bibliographystyle{IEEEtran}
\bibliography{main}
}

\end{document}